%% file: 0872.tex
\documentclass{bmvc2k}

\title{\method: Explicit Structured Query Priors for Temporal 3D Object Detection}

\addauthor{Honglin Sun}{hsun@akane.waseda.jp}{1}
\addauthor{Yifei Zhang}{yifei.zhang@toki.waseda.jp}{1}
\addauthor{Junyang Zhang}{junyang.z@akane.waseda.jp}{1}
\addauthor{Kenji Hashimoto}{kenji.hashimoto@waseda.jp}{1}

\addinstitution{
 Graduate School of Information, Production and Systems\\
 Waseda University\\
 Fukuoka, Japan
}

\runninghead{Sun et al.}{\method}

\newcommand{\method}{ExQuery}

\usepackage{booktabs}
\usepackage{colortbl}
\usepackage{makecell}
\usepackage{adjustbox}
\usepackage{amsmath}
\usepackage{amssymb}
\usepackage{wrapfig}
\usepackage[normalem]{ulem}

\begin{document}
\maketitle

\input{chapters/0-abstract}
\input{chapters/1-introduction}
\input{chapters/2-related-work}
\input{chapters/3-method}
\input{chapters/4-experiments}
\input{chapters/5-conclusion}

\section*{Acknowledgements}
This study was conducted with the support of the Information, Production and Systems Research Center, Waseda University, and the Future Robotics Organization, Waseda University, and was carried out as part of the humanoid project at the Humanoid Robotics Institute, Waseda University. This work was supported by JSPS KAKENHI Grant Number 25K03204. This work was also supported by the Waseda University Grant for Special Research Projects (Project Number: 2026C-179) and by JST BOOST (Grant Number JPMJBS2429).

\bibliography{chapters/references}

\end{document}

%% file: chapters/0-abstract.tex
\begin{abstract}
Sparse query-based 3D detectors commonly encode heterogeneous semantic, geometric, and motion information into unified latent queries, including when explicit 3D anchors are used.
We introduce \method{}, an explicit structured query framework for temporal 3D object detection.
Each decoded query is organized into a shared semantic-context slice and physical-state-specific slices for center, dimensions, orientation, and velocity.
This structure is preserved from prior encoding to branch-specific regression and is reused to reconstruct temporal queries: predicted physical states are transformed by ego motion, classification scores form soft semantic priors, and the reconstructed query is fused with the stored latent instance feature through gated affine modulation.
Experiments on the nuScenes benchmark show that our method achieves competitive detection performance, and ablation studies validate the effectiveness of our proposed modules.
\end{abstract}

%% file: chapters/1-introduction.tex
\section{Introduction}
\label{sec:introduction}

\begin{figure*}[!htbp]
    \centering
    \includegraphics[scale=0.6]{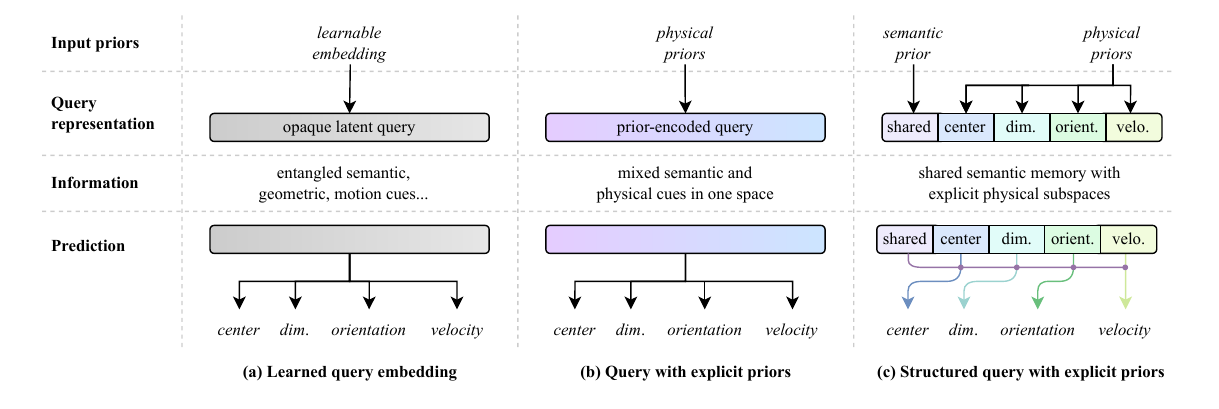}
    \caption{Comparison of different query formulations.
    (a) Learned queries represent object states as opaque latent vectors.
    (b) Prior-encoded queries introduce explicit physical priors but still decode them into a single mixed representation.
    (c) Our structured query state preserves shared semantic-context and physical-state-specific slices for attribute-aware prediction.}
\label{fig:key_idea}
\end{figure*}

3D object detection is a fundamental component of autonomous-driving perception, providing the object category, position, dimensions, orientation, and velocity required by downstream tasks. With the rapid progress of query-based end-to-end detectors since DETR \cite{detr}, vision-based BEV (bird's-eye view) 3D object detection has also advanced significantly. Transformer-based \cite{attention} view transformation avoids hand-designed post-processing and offers a clean formulation for multi-view reasoning. DETR3D \cite{detr3d} directly manipulates sparse 3D object queries, PETR \cite{petr} encodes 3D coordinates into position-aware representations, and PETRv2 \cite{petrv2} introduces temporal image-feature fusion. BEVFormer \cite{bevformer} instead learns dense BEV queries with spatiotemporal Transformers. More recently, StreamPETR \cite{streampetr} and Sparse4D \cite{sparse4dv2,sparse4dv3} have demonstrated the efficiency of instance-centric temporal propagation.


Despite this progress, the role of explicit query priors in 3D detection remains less systematically explored than in 2D DETR-style detectors. 
In 2D detection, query design has evolved from purely learned embeddings to more explicit formulations, such as reference points, dynamic anchor boxes, and anchor boxes with denoising \cite{dabdetr,dndetr,dino}. These designs provide strong geometric guidance for the model and improve detection performance. However, directly extending this explicit-prior paradigm to 3D detection is nontrivial. A 3D object query must represent heterogeneous information, including category, center, dimensions, orientation, velocity, and temporal context. Naively encoding these physical states into a single query space may entangle semantic, geometric, and motion-related factors, making it difficult for different prediction branches to exploit the corresponding physical priors effectively.

Existing 3D detectors have introduced explicit spatial priors such as 3D centers and full 3D anchors. Building on these explicit query representations and instance-centric temporal propagation, our contribution is to preserve an attribute-aligned channel structure in the decoded query and reuse the same structure for temporal reconstruction and branch-specific regression. This organization is maintained in the decoded query itself, providing a consistent representation for constructing and reading out semantic, geometric, and motion states.


These observations motivate a query formulation that not only introduces explicit priors, but also preserves their structure after decoding. 
Simply encoding richer physical priors into a query does not guarantee that the resulting representation remains organized, since semantic, geometric, and motion-related cues may still be mixed in a single latent space.

In this work, we propose \method{}, an explicit structured query framework for sparse query-based 3D object detection.
Instead of representing each object query as an undifferentiated vector, we decompose it into a shared semantic-context component and several physical-state-specific components, including center, dimensions, orientation, and velocity.
As illustrated in Fig.~\ref{fig:key_idea}, this formulation differs from conventional learned queries and prior-encoded queries by maintaining a structured query state after decoding. 
The shared component preserves category-related, contextual, and residual implicit cues, while the physical components provide attribute-aligned subspaces for explicit object states. Based on this structured query state, we further make explicit priors compatible with instance-centric temporal propagation. For stored predictions from previous frames, the physical states are transformed to the current ego frame according to ego-motion, while semantic-context priors are constructed from classification scores. The resulting explicit query is then fused with the stored implicit instance feature through mutual gated affine modulation. In this way, we can maintain a structured propagated query that explicitly carries physical information while still benefiting from contextual latent cues.

The main contributions of this work are as follows:

\begin{itemize}
    \item We propose a structured query formulation that organizes each query into a shared semantic-context component and several explicit physical components. This provides attribute-aligned inputs for branch-specific regression.
    
    \item We introduce an explicit temporal propagation mechanism built on the structured query state, so that propagated queries preserve physical interpretability together with semantic cues across frames. The resulting semantic interface also supports label embeddings during denoising training.
    
    \item Experiments on the nuScenes benchmark show that the proposed method achieves competitive detection performance and validate the effectiveness of structured query modeling for temporal 3D detection.
\end{itemize}

%% file: chapters/2-related-work.tex
\section{Related Work}
\label{sec:related-work}

\subsection{Camera-based 3D Object Detection}

Camera-based 3D object detection has developed along several major lines. Early image-centric methods, such as FCOS3D \cite{fcos3d} and SMOKE \cite{smoke}, directly predict 3D attributes from perspective-view (PV) image features without constructing an explicit BEV representation. While simple and efficient, they are limited by weak multi-view geometric reasoning and less explicit 3D spatial modeling. Subsequent works on multi-view 3D object detection \cite{lss, bevdet, bevdet4d, bevdepth, bevstereo} construct BEV features from images through view transformation with depth estimation. But the dense depth lifting limits the accuracy, temporal modeling, and robustness. Beyond depth-based view transformation, another branch performs BEV reasoning with dense BEV queries using Transformers \cite{bevformer, bevformerv2, polarformer}, eliminating vulnerable depth estimation while making spatiotemporal modeling more natural. In parallel, sparse query-based detectors \cite{detr3d, petr, petrv2, focalpetr, graphdetr3d, sparsebev} have become increasingly attractive due to their object-centric sparse representation and end-to-end detection formulation. Unlike methods that explicitly construct dense BEV queries over the entire scene, they maintain a compact set of sparse object queries, reducing computation on empty BEV regions and improving scalability for multi-view 3D detection.

\subsection{Temporal Modeling for Multi-View 3D Detection}

Temporal information has been incorporated into camera-only 3D detection in several ways.
One line performs temporal fusion in the perspective-view or image domain, where historical image features are aligned and aggregated across frames \cite{petrv2, solofusion}. Another line performs temporal fusion in BEV space \cite{bevformer, bevformerv2, bevdet4d}. These methods benefit from explicit spatial alignment in BEV space, but they also inherit the computation cost of dense BEV representations. OPEN \cite{open} performs temporal aggregation inside an object-wise depth encoder and injects the resulting object-wise position embedding into the decoder. In sparse query-based detection, StreamPETR \cite{streampetr} and QTNet \cite{qtnet} propagate or fuse latent query features using motion-aligned center priors, while Sparse4Dv2 \cite{sparse4dv2} recurrently propagates a latent instance feature together with a full 3D anchor. Compared with image-history or dense BEV-history fusion, instance-centric temporal propagation is particularly attractive for online perception because it preserves the efficiency of sparse detectors. Inspired by the instance-centric propagation of StreamPETR and Sparse4Dv2, our temporal design re-encodes the ego-motion-aligned physical state and score-weighted semantic prior into the structured channel layout used by the decoder and regression heads, then combines it with the stored instance feature through gated fusion.
Similarly, LiDAR-based methods employ motion-aware temporal fusion: MGTANet aligns and aggregates motion-guided BEV features \cite{mgtanet}, whereas MSF propagates motion-aligned proposals and fuses their proposal features \cite{msf}. These methods operate on different input modalities and representation granularities, but provide complementary context for motion-aware temporal modeling.

\subsection{Query Formulation}

Query design in DETR-style detection has gradually evolved from learned queries \cite{detr, deformabledetr} to more explicit spatial priors \cite{anchordetr, conditionaldetr, dabdetr} such as center points or anchor boxes. These designs show that explicit query priors can provide useful geometric guidance for decoder attention and box regression. Denoising-based methods \cite{dndetr, dino} further improve optimization by injecting noised label and box queries during training. In 3D object detection, query formulation has evolved less uniformly. Early Transformer-based methods \cite{detr3d, bevformer} largely follow learned-query formulations, while later sparse-query detectors introduce explicit priors through learnable 3D center points \cite{petr, petrv2, focalpetr, streampetr}, richer 3D anchor boxes \cite{sparse4dv2, sparse4dv3}, or 2D-guided query generation \cite{mv2d, far3d, qaf2d}. Compared with 2D detection, a 3D query must represent more heterogeneous information, including semantic cues, physical states, and context information. Encoding them into a single query space may entangle different information, making explicit query-prior design in 3D nontrivial. Related decoupling strategies have been explored in both 2D and 3D detection. In 2D DETR literature, classification and localization decoupling has been studied to reduce interference between semantic and spatial information \cite{destr, decoupleddetr}. In 3D detection, Sparse4D v3 \cite{sparse4dv3} adopts decoupled attention to alleviate feature interference between anchor embeddings and instance features. Our work focuses on a structured decoded query with a shared semantic-context component and physical-state-specific components for different box attributes, and preserves this organization from prior encoding through temporal reconstruction to branch-specific regression.

%% file: chapters/3-method.tex
\section{Method}
\label{sec:method}

\subsection{Overview}
\label{sec:overview}

\begin{figure*}[!htbp]
    \centering
    \includegraphics[scale=0.7]{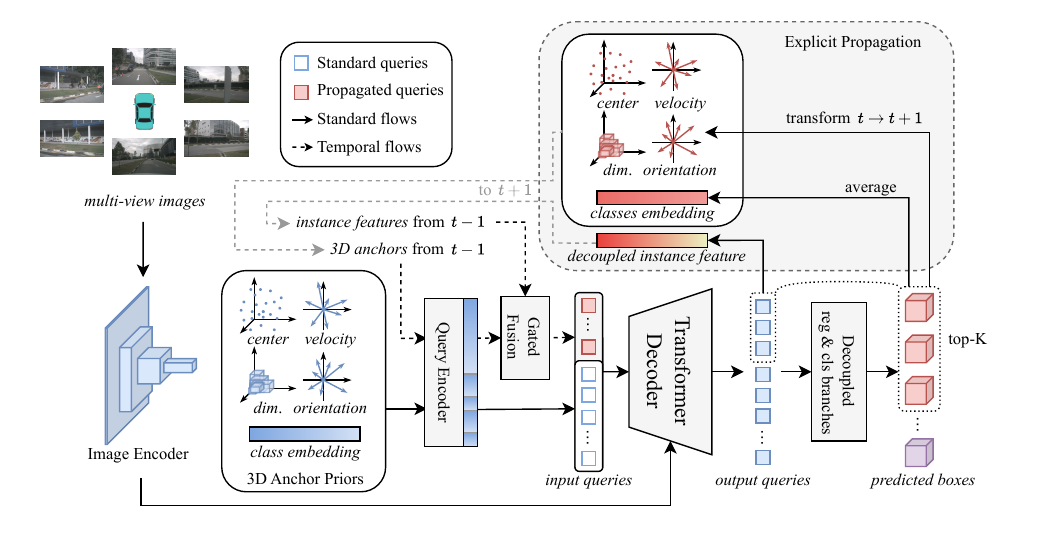}
    \caption{Overview of the proposed framework. Multi-view images are encoded into image features, while learnable 3D anchor priors, consisting of explicit physical states and class embeddings, are encoded into input queries. The Transformer decoder takes both newly encoded queries and propagated queries to predict 3D bounding boxes. At each time step, the top-$K$ predictions together with their corresponding instance features are stored in a memory queue and propagated to the next frame, enabling structured query-state propagation with explicit physical states and semantic-context priors.}
\label{fig:framework}
\end{figure*}

The overall framework of our method is illustrated in Fig. \ref{fig:framework}. At each time step $t$, we take multi-view images $\{\mathbf{I}_i^t\}_{i=1}^{N}$ as input and extract multi-scale image features $\{\mathbf{F}_{i,l}^t\}_{i=1,l=1}^{N,L}$ using an image backbone (\textit{e.g.}, ResNet \cite{resnet}, VoVNet \cite{vovnet}) and an image neck (\textit{e.g.}, FPN \cite{fpn}).

We maintain a set of learnable 3D anchors, where each anchor is associated with an explicit physical state
\begin{equation}
    \mathbf{a} = \left[x, y, z, w, l, h, \theta_x, \theta_y, v_x, v_y \right],
\end{equation}
including object center, dimensions, orientation, and velocity. In addition, a learnable neutral class embedding is introduced as the initial semantic prior. These anchor priors are encoded by a query encoder to form the input queries of the Transformer decoder.

Given the image features and encoded input queries, the Transformer decoder updates the queries through multi-head attention \cite{attention}, where the queries aggregate relevant information from multi-view image features. The output queries are fed into the classification and regression heads to predict object categories and 3D bounding boxes. To maintain a structured representation, we organize each query into a shared semantic-context slice and several physical-state-specific slices, as detailed in Sec. \ref{sec:structured-query-state}.

For temporal propagation, we reconstruct each explicit query state from a predicted physical state and a semantic-context prior, while the corresponding stored decoder instance feature provides complementary implicit information.

At time step $t$, we store the top-$K$ predictions together with their instance features in a memory queue. At the next time step $t+1$, the propagated physical states are first transformed according to the ego-motion between frames. The transformed physical states and the propagated semantic-context priors are then encoded by the query encoder to form the raw propagated temporal queries. On top of this, we further use the stored instance features as complementary implicit cues to modulate the propagated queries through a gated fusion module. The resulting propagated queries are then used together with the current-frame image features. In this way, our framework achieves propagation with an explicit scheme.

\subsection{Structured Query State}
\label{sec:structured-query-state}

Instead of representing each object query as an undifferentiated latent vector, we model it as a structured query state. Specifically, the feature channels are divided into a shared semantic-context slice and several physical-state-specific slices:
\begin{equation}
    \mathbf{q}
    =
    \left[
    \mathbf{q}_s
    \mid
    \mathbf{q}_c
    \mid
    \mathbf{q}_d
    \mid
    \mathbf{q}_o
    \mid
    \mathbf{q}_v
    \right],
\end{equation}
where $\mathbf{q}_s$ denotes the shared semantic-context slice, and $\mathbf{q}_c$, $\mathbf{q}_d$, $\mathbf{q}_o$, and $\mathbf{q}_v$ correspond to center, dimensions, orientation, and velocity, respectively. The shared slice preserves implicit cues such as semantic, contextual, and temporal information, while the physical slices are aligned with explicit object states. This design provides attribute-aligned subspaces for temporal propagation and regression.

Given the 3D anchor state defined in Sec. \ref{sec:overview}, the query encoder maps its semantic and physical priors into the query feature space. The semantic prior is a learnable embedding for standard matching queries and is replaced by a propagated semantic-context prior for temporal queries. Center and dimensions are encoded with sinusoidal positional encodings followed by lightweight projections. For orientation and velocity, we use a complex-valued positional encoding (CPE) to preserve their directional structure.

For a two-dimensional vector $\mathbf{u}=(u_1,u_2)$, we first map it to a complex directional representation
\begin{equation}
    z(\mathbf{u})
    =
    \begin{cases}
        \dfrac{u_1 + j u_2}{\|\mathbf{u}\|_2}, & \|\mathbf{u}\|_2 > 0,\\[4pt]
        0, & \|\mathbf{u}\|_2 = 0,
    \end{cases}
\end{equation}
where $j$ is the imaginary unit.
Given $F$ angular frequencies, the CPE is defined as
\begin{equation}
    \mathrm{CPE}(\mathbf{u})
    =
    \left[
    \operatorname{Re}\left(z(\mathbf{u})^{2^i}\right),
    \operatorname{Im}\left(z(\mathbf{u})^{2^i}\right)
    \right]_{i=0}^{F-1}
    \in \mathbb{R}^{2F},
\end{equation}
where $\operatorname{Re}(\cdot)$ and $\operatorname{Im}(\cdot)$ denote the real and imaginary parts, respectively, and $[\cdot]_{i=0}^{F-1}$ denotes collecting the features over all frequencies into a single vector.

This encoding follows the periodic structure of direction vectors and provides multi-frequency angular features. Orientation uses this directional code directly. Velocity additionally retains speed information. Let $\bar r_v$ denote the speed $\|\mathbf a_v\|_2$ normalized by the configured velocity range, and let $\mathrm{SPE}(\bar r_v)$ be its sinusoidal magnitude encoding. The orientation and velocity embeddings are
\begin{equation}
    \mathbf{e}_o
    =
    \mathcal{P}_o
    \left(
    \mathrm{CPE}(\mathbf{a}_o)
    \right),
    \qquad
    \mathbf{e}_v
    =
    \mathcal{P}_v
    \left(
    [\mathrm{CPE}(\mathbf{a}_v)\mid\mathrm{SPE}(\bar r_v)]
    \right),
\end{equation}
where $\mathbf{a}_o$ and $\mathbf{a}_v$ denote the orientation and velocity priors, respectively. For zero velocity, the directional code is the zero vector, whereas the magnitude branch explicitly represents the zero-speed state.

The encoded semantic and physical components are then fused by the structured query encoder $\mathcal{G}$ and reallocated into the slice-aligned query channels defined above. The same channel structure is maintained after decoding. For box regression, the branch associated with each physical attribute takes the shared slice together with its corresponding physical-state-specific slice:
\begin{equation}
    \hat{\mathbf{y}}_k
    =
    f_k
    \left(
    [\mathbf{q}_s \mid \mathbf{q}_k]
    \right),
    \qquad
    k\in\{c,d,o,v\}.
\end{equation}
Here $\hat{\mathbf{y}}_c$, $\hat{\mathbf{y}}_d$, $\hat{\mathbf{y}}_o$, and $\hat{\mathbf{y}}_v$ denote the predicted center, dimensions, orientation, and velocity, respectively. For classification, we use the full decoder query, since category prediction in 3D detection can still benefit from correlated geometry and motion cues.




\subsection{Explicit Temporal Propagation}
\label{sec:explicit-temporal-propagation}

At each time step $t$, we select the top-$K$ high-confidence predictions and store their predicted physical states, classification scores, and decoder output instance features in a memory queue. The propagated query is thus constructed from an explicitly encoded query state and a complementary implicit instance feature.

For each stored instance, the predicted physical state is first transformed to the current ego frame. Let $\hat{\mathbf{a}}_i^{t}$ denote the predicted physical state of the $i$-th stored instance at time $t$. Its propagated physical prior at time $t+1$ is
\begin{equation}
\begin{aligned}
    \mathbf{a}_i^{t+1}
    &=
    \mathcal{T}_{t}^{t+1}\!\left(\hat{\mathbf{a}}_i^{t}\right),
    &\mathbf{T}_{t}^{t+1}&=\mathbf{T}_{t+1}^{-1}\mathbf{T}_{t},\\
    \mathbf{T}_{t}
    &=
    \begin{bmatrix}
        \mathbf{R}_{t} & \mathbf{p}_{t}\\
        \mathbf{0} & 1
    \end{bmatrix},
\end{aligned}
\end{equation}
where $\mathbf{T}_{t}$ is the ego-to-global transformation at time $t$, with rotation $\mathbf R_t$ and translation $\mathbf p_t$. The induced operation $\mathcal T_t^{t+1}$ transforms box centers by $\mathbf T_t^{t+1}$, leaves dimensions unchanged, and applies its relative rotation to orientation and velocity.

We also construct a semantic-context prior from the classification scores. Let $\mathbf{p}_i^t \in [0,1]^{C}$ be the class confidence scores of the $i$-th stored prediction, and let $\mathbf{E}=\{\mathbf{e}_c\}_{c=1}^{C}$ be the learnable class embedding table shared with the denoising branch. We select the top-$M$ classes $\mathcal{C}_i^M$ and compute a score-weighted class embedding:
\begin{equation}
    \bar{\mathbf{e}}_i^t
    =
    \sum_{c\in \mathcal{C}_i^M}
    \alpha_{i,c} \mathbf{e}_c,
    \qquad
    \alpha_{i,c}
    =
    \frac{p_{i,c}^t}
    {\sum_{c'\in \mathcal{C}_i^M} p_{i,c'}^t}.
\end{equation}
Here, $\alpha_{i,c}$ denotes the normalized confidence weight of class $c$ among the selected top-$M$ classes, and $\bar{\mathbf e}_i^t$ serves as a soft semantic prior for the stored instance.

To account for low-confidence stored predictions, we mix this class-aware prior with a learnable neutral embedding $\mathbf{e}_{0}$:
\begin{equation}
    p_i^{\mathrm{obj}}=\max_{c} p_{i,c}^t,
    \qquad
    \mathbf{e}_{i}^{t+1}
    =
    p_i^{\mathrm{obj}} \bar{\mathbf{e}}_i^t
    +
    \left(1-p_i^{\mathrm{obj}}\right)\mathbf{e}_{0}.
\end{equation}

The transformed physical prior $\mathbf{a}_i^{t+1}$ and semantic-context prior $\mathbf{e}_{i}^{t+1}$ are then encoded as an explicit structured query:
\begin{equation}
    \mathbf{q}_{\mathrm{exp},i}^{t+1}
    =
    \phi_{\mathrm{enc}}
    \left(
    \mathbf{e}_{i}^{t+1},
    \mathbf{a}_i^{t+1}
    \right).
\end{equation}

\begin{figure*}[!htbp]
    \centering
    \hspace*{12mm}
    \includegraphics[scale=0.9]{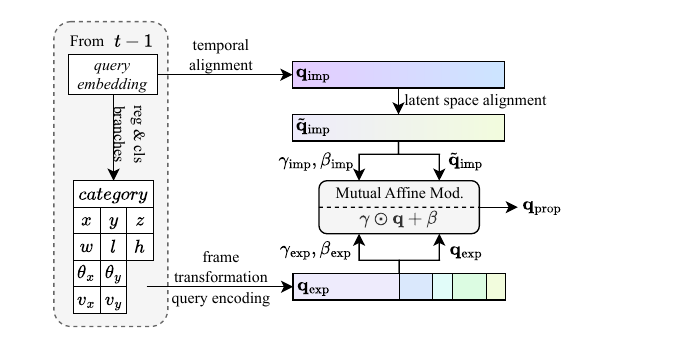}
    \caption{
    Mutual gated affine fusion for temporal query propagation.
    The score-derived semantic prior and predicted physical state form the explicit structured query $\mathbf q_{\mathrm{exp}}$, while the stored decoder feature forms the implicit branch $\mathbf q_{\mathrm{imp}}$.
}
\label{fig:gate}
\end{figure*}

As illustrated in Fig.\ref{fig:gate}, we further fuse this explicit structured query with the stored implicit instance feature through mutual gated affine modulation.

To preserve useful implicit contextual information from the previous frame, we further modulate the explicit query with the stored instance feature. Let $\mathbf{q}_{\mathrm{imp},i}^{t}$ denote the stored instance feature. We first project it to the query space:
\begin{equation}
    \tilde{\mathbf{q}}_{\mathrm{imp},i}^{t}
    =
    \phi_{\mathrm{align}}
    \left(
    \mathbf{q}_{\mathrm{imp},i}^{t}
    \right).
\end{equation}
We then perform mutual gated affine fusion:
\begin{equation}
    \boldsymbol{\gamma}_{\mathrm{imp}}, \boldsymbol{\beta}_{\mathrm{imp}}
    =
    \phi_{\mathrm{imp}}
    \left(
    \tilde{\mathbf{q}}_{\mathrm{imp},i}^{t}
    \right),
    \qquad
    \boldsymbol{\gamma}_{\mathrm{exp}}, \boldsymbol{\beta}_{\mathrm{exp}}
    =
    \phi_{\mathrm{exp}}
    \left(
    \mathbf{q}_{\mathrm{exp},i}^{t+1}
    \right),
\end{equation}
\begin{equation}
    \mathbf{q}_{\mathrm{prop},i}^{t+1}
    =
    \boldsymbol{\gamma}_{\mathrm{imp}}
    \odot
    \mathbf{q}_{\mathrm{exp},i}^{t+1}
    +
    \boldsymbol{\beta}_{\mathrm{imp}}
    +
    \boldsymbol{\gamma}_{\mathrm{exp}}
    \odot
    \tilde{\mathbf{q}}_{\mathrm{imp},i}^{t}
    +
    \boldsymbol{\beta}_{\mathrm{exp}},
\end{equation}
where $\phi$ denotes an MLP, $\boldsymbol{\gamma}$ and $\boldsymbol{\beta}$ are channel-wise affine parameters, and $\odot$ denotes element-wise multiplication.

Finally, the propagated queries are concatenated with the newly initialized standard queries at time $t+1$ and fed into the Transformer decoder together with the current-frame image features.



%% file: chapters/4-experiments.tex
\section{Experiments}
\label{sec:experiments}

\subsection{Dataset and Metrics}

We conduct experiments on the nuScenes dataset \cite{nuscenes}, a large-scale autonomous driving benchmark containing 1000 driving scenes of approximately 20 seconds each. The dataset is officially split into 700 scenes for training, 150 scenes for validation, and 150 scenes for testing. We train our models on the training set and report results on the validation set.

We evaluate our method with the official nuScenes 3D detection metrics, including mean Average Precision (mAP), nuScenes Detection Score (NDS), and five true-positive error metrics: mean Average Translation Error (mATE), mean Average Scale Error (mASE), mean Average Orientation Error (mAOE), mean Average Velocity Error (mAVE), and mean Average Attribute Error (mAAE). The mAP is computed based on center-distance matching on the ground plane, while NDS summarizes both detection accuracy and box quality. Higher mAP and NDS indicate better performance, whereas lower true-positive errors indicate more accurate localization, scale, orientation, velocity, and attribute estimation.


\subsection{Implementation Details}

Our implementation is based on RepDETR3D, the deformable-attention variant included in the official StreamPETR \cite{streampetr} codebase, with its original MMCV1 implementation adapted to an MMEngine-based environment.

Following StreamPETR, we use a ResNet-50 \cite{resnet} backbone initialized from a nuImages-pretrained Cascade R-CNN, an FPN \cite{fpn} neck, $256\times704$ input images, 300 standard queries, and 128 propagation queries. Models are trained on NVIDIA RTX 5090 GPUs with a total batch size of 16. We use AdamW \cite{adamw}, cosine learning-rate scheduling, and an initial learning rate of $4\times10^{-4}$. The main model is trained for 60 epochs without class-balanced group sampling (CBGS) \cite{cbgs}, while ablation models use 24 epochs under the same protocol.


\subsection{Main Results}


\begin{table}[t]
\centering
\resizebox{\linewidth}{!}{
\begin{tabular}{@{}lccccccccc@{}}
\toprule
\multicolumn{1}{l|}{\textbf{Method}}            & \multicolumn{1}{c|}{\textbf{Backbone}} & \multicolumn{1}{c|}{\textbf{Resolution}} & \textbf{NDS$\uparrow$} & \multicolumn{1}{c|}{\textbf{mAP$\uparrow$}} & \multicolumn{1}{c}{\textbf{mATE$\downarrow$}} & \textbf{mASE$\downarrow$} & \textbf{mAOE$\downarrow$} & \textbf{mAVE$\downarrow$} & \textbf{mAAE$\downarrow$} \\ 
\midrule
\multicolumn{1}{l|}{PETRv2\cite{petrv2}}        & \multicolumn{1}{c|}{ResNet50}  & \multicolumn{1}{c|}{256 $\times$ 704}  & 0.456 & \multicolumn{1}{c|}{0.349} & 0.700 & 0.275 & 0.580 & 0.437 & 0.187 \\
\multicolumn{1}{l|}{BEVDet4D\cite{bevdet4d}}    & \multicolumn{1}{c|}{ResNet50}  & \multicolumn{1}{c|}{256 $\times$ 704}  & 0.457 & \multicolumn{1}{c|}{0.322} & 0.703 & 0.278 & 0.495 & 0.354 & 0.206 \\
\multicolumn{1}{l|}{BEVDepth\cite{bevdepth}}    & \multicolumn{1}{c|}{ResNet50}  & \multicolumn{1}{c|}{256 $\times$ 704}  & 0.475 & \multicolumn{1}{c|}{0.351} & 0.639 & 0.267 & 0.479 & 0.428 & 0.198 \\
\multicolumn{1}{l|}{BEVStereo\cite{bevstereo}}  & \multicolumn{1}{c|}{ResNet50}  & \multicolumn{1}{c|}{256 $\times$ 704}  & 0.500 & \multicolumn{1}{c|}{0.372} & 0.598 & 0.270 & 0.438 & 0.367 & 0.190 \\
\multicolumn{1}{l|}{SOLOFusion\cite{solofusion}}& \multicolumn{1}{c|}{ResNet50}  & \multicolumn{1}{c|}{256 $\times$ 704}  & 0.534 & \multicolumn{1}{c|}{0.427} & 0.567 & 0.274 & 0.511 & 0.252 & 0.181 \\
\multicolumn{1}{l|}{Sparse4Dv2\cite{sparse4dv2}}& \multicolumn{1}{c|}{ResNet50}  & \multicolumn{1}{c|}{256 $\times$ 704}  & 0.539 & \multicolumn{1}{c|}{0.439} & 0.598 & 0.270 & 0.475 & 0.282 & \textbf{0.179} \\
\multicolumn{1}{l|}{StreamPETR$^\dagger$\cite{streampetr}}& \multicolumn{1}{c|}{ResNet50}  & \multicolumn{1}{c|}{256 $\times$ 704}  & 0.550 & \multicolumn{1}{c|}{0.450} & 0.613 & 0.267 & 0.413 & 0.265 & 0.196 \\
\multicolumn{1}{l|}{SparseBEV$^\dagger$\cite{sparsebev}}& \multicolumn{1}{c|}{ResNet50}  & \multicolumn{1}{c|}{256 $\times$ 704}  & 0.558 & \multicolumn{1}{c|}{0.448} & 0.581 & 0.271 & \textbf{0.373} & \textbf{0.247} & 0.190 \\
\multicolumn{1}{l|}{BEVNeXt$^\dagger$\cite{bevnext}}& \multicolumn{1}{c|}{ResNet50}  & \multicolumn{1}{c|}{256 $\times$ 704}  & 0.560 & \multicolumn{1}{c|}{0.456} & \textbf{0.530} & 0.264 & 0.424 & 0.252 & 0.206 \\
\rowcolor[HTML]{ECECEC}
\multicolumn{1}{l|}{\textbf{\method{}}$^\dagger$}  & \multicolumn{1}{c|}{ResNet50}  & \multicolumn{1}{c|}{256 $\times$ 704}  & \textbf{0.561} & \multicolumn{1}{c|}{\textbf{0.459}} & 0.585 & \textbf{0.261} & 0.400 & 0.256 & 0.193 \\
\bottomrule
\end{tabular}
}
\caption{Comparison results on the nuScenes validation set. $\dagger$ denotes perspective-view pretraining.}
\label{tab:nuscenes_val}
\end{table}

Table~\ref{tab:nuscenes_val} compares our method with representative camera-based 3D object detectors on the nuScenes validation set. With the ResNet-50 image backbone and an input resolution of $256\times704$, our method achieves 0.459 mAP and 0.561 NDS. Compared with StreamPETR, our method improves mAP from 0.450 to 0.459 and NDS from 0.550 to 0.561. With the explicit propagated priors, it also reduces several box-quality errors, including mATE, mASE, mAOE, and mAVE, indicating that the proposed method benefits detection accuracy.

\begin{table}[t]
\centering
\small
\setlength{\tabcolsep}{4pt}
\begin{tabular}{lccccc}
\toprule
\textbf{Model} & \textbf{Params.} & \textbf{GFLOPs} & \textbf{Mem.} & \textbf{FPS} & \textbf{Latency} \\
 & (M) & & (MB) & & (ms) \\
\midrule
RepDETR3D & 43.29 & 225.03 & 404.12 & 67.22 & 14.88 \\
\method   & 46.17 & 227.60 & 415.69 & 54.10 & 18.48 \\
\bottomrule
\end{tabular}
\caption{FP32 efficiency of the RepDETR3D base and \method{} at batch size 1 on an RTX 5090.}
\label{tab:efficiency}
\end{table}

RepDETR3D and \method{} use the same ResNet-50 backbone, FPN neck, $256\times704$ input resolution, and runtime settings. Table~\ref{tab:efficiency} reports the computational overhead of \method{} under this matched setup: 2.88M parameters, 2.57 GFLOPs, and 11.57 MB of measured memory, with latency increasing from 14.88 to 18.48 ms.

\subsection{Ablation Study}

\begin{table}[ht]
\centering
\small
\setlength{\tabcolsep}{6pt}
\begin{tabular}{c c c|c c}
\toprule
\textbf{Structured Query} & \textbf{Explicit Prior} & \textbf{Gated Fusion} & \textbf{NDS$\uparrow$} & \textbf{mAP$\uparrow$}   \\
\midrule
                 &                &              & 0.470    & 0.383 \\
\checkmark       &                &              & 0.497    & 0.386 \\
                 & \checkmark     &              & 0.501    & 0.391 \\
\checkmark       & \checkmark     &              & 0.516    & 0.406 \\
\rowcolor[HTML]{ECECEC}
\checkmark       & \checkmark     & \checkmark   & \textbf{0.522}    & \textbf{0.417} \\
\bottomrule
\end{tabular}
\caption{Ablation study of the proposed components on the nuScenes validation set.}
\label{tab:component_ablation}
\end{table}

\textbf{Effect of main components.}
Table \ref{tab:component_ablation} studies the contribution of the proposed components. Here, Structured Query denotes the channel-sliced query representation with decoupled regression heads. Explicit Prior denotes the explicitly propagated physical and semantic-context priors. Gated Fusion denotes the mutual gated affine fusion between the explicit query and the stored implicit instance feature. Introducing the structured query improves NDS from 0.470 to 0.497 and mAP from 0.383 to 0.386, indicating that organizing the query into attribute-aligned subspaces benefits 3D detection. Adding the explicit prior further improves the performance to 0.516 NDS and 0.406 mAP, showing the benefit of explicitly propagating physical and semantic-context information across frames. With mutual gated affine fusion, the full model achieves 0.522 NDS and 0.417 mAP, suggesting that the structured query, explicit prior, and gated fusion are complementary.

\begin{table*}[ht]
\centering
\small
\setlength{\tabcolsep}{4pt}

\begin{minipage}[t]{0.48\linewidth}
\vspace{0pt}
\centering
\begin{tabular}{c c c|c c}
\toprule
$\mathbf{q}_{\mathrm{exp}}$ & $\mathbf{q}_{\mathrm{imp}}$   & \textbf{Fusion} & \textbf{NDS$\uparrow$} & \textbf{mAP$\uparrow$} \\
\midrule
                             & \checkmark                   & --                      & 0.487 & 0.378 \\
\checkmark                   &                              & --                      & 0.501 & 0.391 \\
\checkmark                   & \checkmark                   & Exp. $\rightarrow$ Imp. & 0.512 & \textbf{0.418} \\
\checkmark                   & \checkmark                   & Exp. $\leftarrow$ Imp.  & 0.517 & 0.414 \\
\rowcolor[HTML]{ECECEC}
\checkmark                   & \checkmark                   & Mutual                  & \textbf{0.522} & 0.417 \\
\bottomrule
\end{tabular}
\caption{Ablation study of propagated query fusion strategies.}
\label{tab:fusion_ablation}
\end{minipage}%
\hfill%
\begin{minipage}[t]{0.48\linewidth}
\vspace{0pt}
\centering
\begin{tabular}{c c|c c}
\toprule
\makecell{\textbf{Query} \\ \textbf{Formulation}} & \makecell{\textbf{Regression} \\ \textbf{Input}} & \textbf{NDS$\uparrow$} & \textbf{mAP$\uparrow$} \\
\midrule
Unified & $\mathbf{q}$                                      & 0.518 & 0.416 \\
Sliced  & $\mathbf{q}$                                      & 0.508 & 0.400 \\
Sliced  & $[\mathbf{0}\mid\mathbf{q}_{k}]$                  & 0.506 & 0.405 \\
\rowcolor[HTML]{ECECEC}
Sliced  & $[\mathbf{q}_{s}\mid\mathbf{q}_{k}]$              & \textbf{0.522} & \textbf{0.417} \\
\bottomrule
\end{tabular}
\caption{Ablation study of query formulation and regression head design.}
\label{tab:structured_query_ablation}
\end{minipage}

\end{table*}

\textbf{Effect of propagated query fusion.}
Table~\ref{tab:fusion_ablation} compares different strategies for constructing propagated queries. Here, $\mathbf{q}_{\mathrm{exp}}$ denotes the explicitly encoded propagated query, while $\mathbf{q}_{\mathrm{imp}}$ denotes the stored instance feature. In the Fusion column, Exp. $\rightarrow$ Imp. means that the explicit query modulates the implicit feature, while Exp. $\leftarrow$ Imp. denotes the reverse direction. Directly propagating only the implicit instance feature achieves 0.487 NDS and 0.378 mAP. Using only the explicitly encoded query improves the performance to 0.501 NDS and 0.391 mAP. This demonstrates the effectiveness of the explicitly encoded physical and semantic-context priors. Mutual gated affine fusion further improves the result to 0.522 NDS and 0.417 mAP, confirming that the stored instance feature still provides useful complementary cues when fused with the explicit query.

\textbf{Effect of structured query design.}
Table~\ref{tab:structured_query_ablation} studies the query formulation and regression input design. 
Unified denotes an unstructured query without channel slicing, while Sliced denotes the proposed shared/physical slice formulation. 
In the Regression Input column, $\mathbf{q}$ denotes the whole query, $\left[\mathbf{0}\mid\mathbf{q}_{k}\right]$ denotes using the corresponding physical-state-specific slice with the shared slice filled with zeros, and $[\mathbf{q}_{s}\mid\mathbf{q}_{k}]$ denotes using the shared slice together with the corresponding physical slice. The results show that simply slicing the query while still using the full query or using only the physical slice for regression does not improve performance, indicating the importance of shared semantic-context information. 
Compared with the unified query, the final sliced query with $[\mathbf{q}_{s}\mid\mathbf{q}_{k}]$ improves NDS from 0.518 to 0.522, while mAP only changes from 0.416 to 0.417. Since NDS accounts for both mAP and 3D box quality metrics in nuScenes, the larger gain in NDS suggests that the physical components improve localization and box quality rather than mainly improving mAP.
This supports the role of the structured query as an interpretable state representation, where each regression branch can access shared semantic-context information together with its corresponding physical-state-specific subspace.

\subsection{Representation Analysis and Visualization}

\textbf{Structured query slice intervention.} To analyze the functional role of each physical slice, we zero out one slice at a time before the prediction heads and measure the metric difference
$\Delta m = m_{\mathrm{intervention}} - m_{\mathrm{normal}}$.
For NDS and mAP, negative values indicate degraded detection performance, while for mATE, mASE, mAOE, and mAVE, positive values indicate increased error. Since the shared slice is designed as a common semantic-context pathway rather than a physical-state-specific slice, zeroing the shared slice causes large degradation across all metrics, so we focus this analysis on the physical slices. 

\begin{figure*}[!htbp]
    \centering
    \includegraphics[scale=0.55]{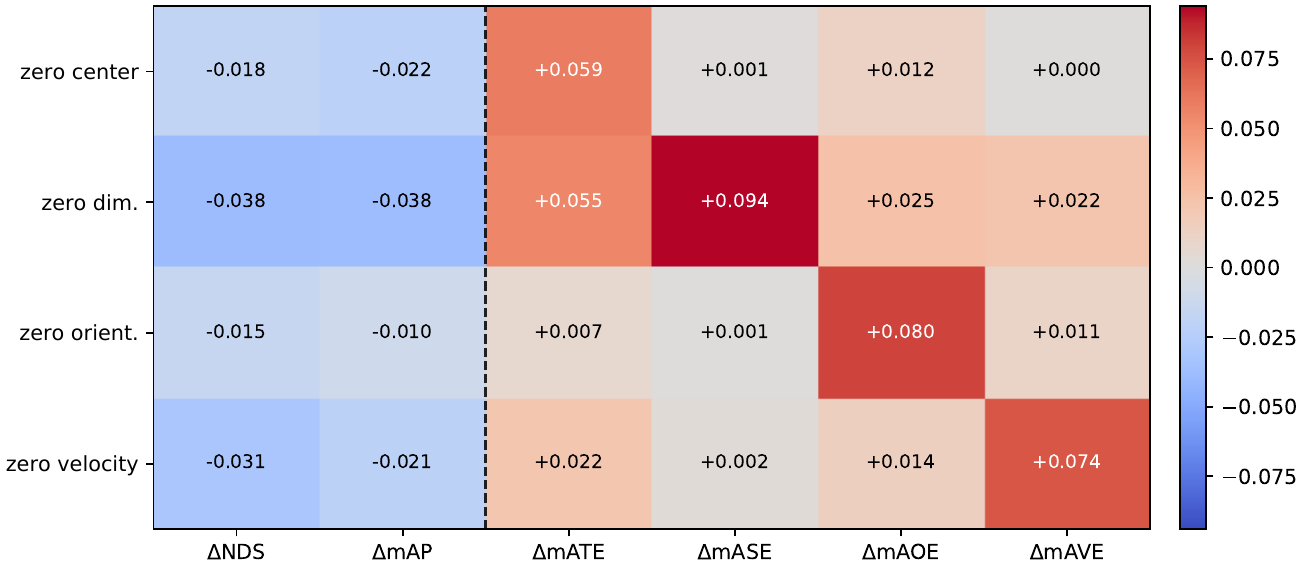}
    \caption{Structured query physical-slice intervention. We zero out each physical slice before the prediction heads and report the signed metric difference. The dashed line separates detection metrics from box-quality error metrics. This intervention probes functional attribute alignment of the physical slices.}
    \label{fig:slice-intervention}
\end{figure*}

As shown in Fig.~\ref{fig:slice-intervention}, zeroing the dimensions, orientation, and velocity slices most strongly affects mASE, mAOE, and mAVE, respectively. The center slice has a less isolated effect, consistent with center estimation also using the shared context. Although these interventions do not establish complete disentanglement, they support attribute alignment of the physical slices.

\begin{figure*}[!htbp]
    \centering
    \includegraphics[scale=0.168]{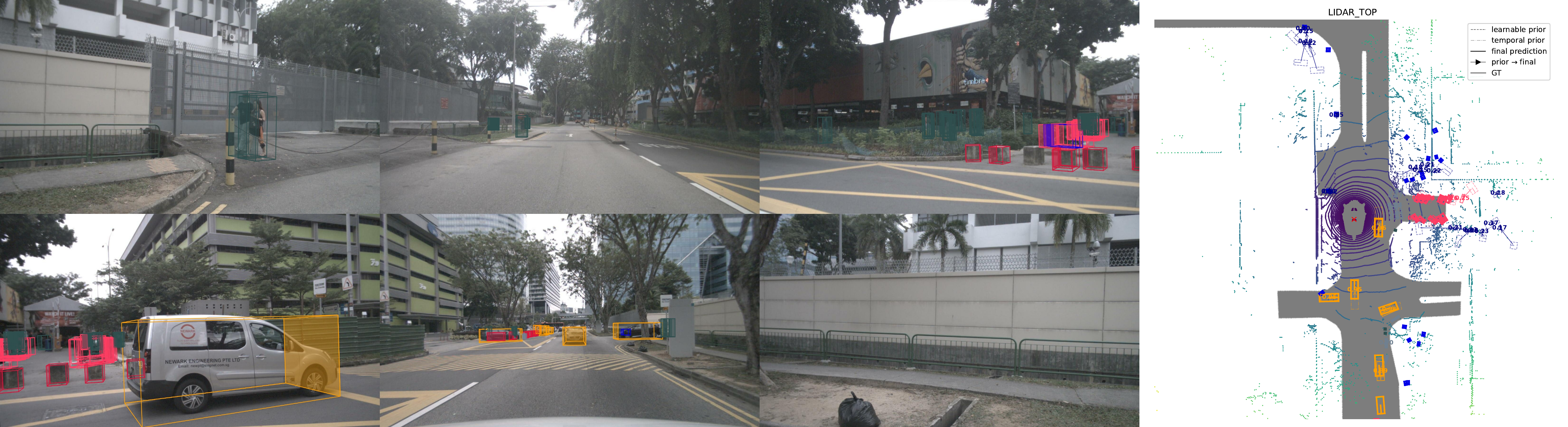}
    \caption{Visualization of detection results.
    In the BEV plot, dashed and dash-dotted boxes denote the learnable prior and temporal prior, respectively.
    Faded boxes show their initial states before decoder refinement, arrows indicate the displacement to final predictions, and thick solid boxes denote ground truth.}
    \label{fig:pred}
\end{figure*}

\textbf{Visualization of Detection Results.} Fig.\ref{fig:pred} visualizes the detection results together with the propagated priors. 
The classification scores are annotated in the BEV plot. In this example, temporal-prior predictions generally obtain higher confidence scores and are closely aligned with the final predictions after decoder refinement. This suggests that the temporal prior provides a strong initialization for propagated objects, reducing the refinement distance required by the decoder. The remaining displacement from the prior to the final prediction mainly reflects object motion and localization refinement. Meanwhile, learnable priors still serve as complementary queries, helping detect objects that are not well covered by temporal propagation or that newly appear in the current frame.

%% file: chapters/5-conclusion.tex
\section{Conclusion and Limitations}

We presented \method{}, which preserves a shared/physical channel organization in 3D object queries from explicit prior encoding to temporal reconstruction and branch-specific regression. Experiments show competitive detection performance, while the ablations support the complementary roles of structured queries, explicit priors, and gated fusion.

Despite these encouraging results, several directions remain open for further exploration. In our current design, the shared semantic-context slice is intentionally used by all prediction branches, allowing physical prediction to benefit from contextual cues. The added query encoders and fusion module also introduce a measurable latency cost. A deeper understanding of what information is involved in the query, and how to further encourage attribute-specific specialization without losing useful contextual interaction, remains an important direction for future work. We hope this work can encourage further study of explicit, structured, and controllable query representations for 3D perception.

%% file: 0872.bbl
\begin{thebibliography}{44}
\providecommand{\natexlab}[1]{#1}
\providecommand{\url}[1]{\texttt{#1}}
\expandafter\ifx\csname urlstyle\endcsname\relax
  \providecommand{\doi}[1]{doi: #1}\else
  \providecommand{\doi}{doi: \begingroup \urlstyle{rm}\Url}\fi

\bibitem[Caesar et~al.(2020)Caesar, Bankiti, Lang, Vora, Liong, Xu, Krishnan,
  Pan, Baldan, and Beijbom]{nuscenes}
Holger Caesar, Varun Bankiti, Alex~H Lang, Sourabh Vora, Venice~Erin Liong,
  Qiang Xu, Anush Krishnan, Yu~Pan, Giancarlo Baldan, and Oscar Beijbom.
\newblock {nuScenes}: A multimodal dataset for autonomous driving.
\newblock In \emph{Proceedings of the IEEE/CVF Conference on Computer Vision
  and Pattern Recognition}, pages 11621--11631, 2020.

\bibitem[Carion et~al.(2020)Carion, Massa, Synnaeve, Usunier, Kirillov, and
  Zagoruyko]{detr}
Nicolas Carion, Francisco Massa, Gabriel Synnaeve, Nicolas Usunier, Alexander
  Kirillov, and Sergey Zagoruyko.
\newblock End-to-end object detection with transformers.
\newblock In \emph{European Conference on Computer Vision}, pages 213--229.
  Springer, 2020.

\bibitem[Chen et~al.(2022)Chen, Li, Zhang, Fang, Jiang, and Zhao]{graphdetr3d}
Zehui Chen, Zhenyu Li, Shiquan Zhang, Liangji Fang, Qinhong Jiang, and Feng
  Zhao.
\newblock {Graph-DETR3D}: Rethinking overlapping regions for multi-view {3D}
  object detection.
\newblock In \emph{Proceedings of the 30th ACM International Conference on
  Multimedia}, pages 5999--6008, 2022.

\bibitem[He et~al.(2023)He, Li, Zhang, Li, and Zhang]{msf}
Chenhang He, Ruihuang Li, Yabin Zhang, Shuai Li, and Lei Zhang.
\newblock {MSF}: Motion-guided sequential fusion for efficient {3D} object
  detection from point cloud sequences.
\newblock In \emph{Proceedings of the IEEE/CVF Conference on Computer Vision
  and Pattern Recognition}, pages 5196--5205, 2023.

\bibitem[He et~al.(2016)He, Zhang, Ren, and Sun]{resnet}
Kaiming He, Xiangyu Zhang, Shaoqing Ren, and Jian Sun.
\newblock Deep residual learning for image recognition.
\newblock In \emph{Proceedings of the IEEE Conference on Computer Vision and
  Pattern Recognition}, pages 770--778, 2016.

\bibitem[He and Todorovic(2022)]{destr}
Liqiang He and Sinisa Todorovic.
\newblock {DESTR}: Object detection with split transformer.
\newblock In \emph{Proceedings of the IEEE/CVF Conference on Computer Vision
  and Pattern Recognition}, pages 9377--9386, 2022.

\bibitem[Hou et~al.(2023)Hou, Liu, Liang, Zou, Ye, and Bai]{qtnet}
Jinghua Hou, Zhe Liu, Dingkang Liang, Zhikang Zou, Xiaoqing Ye, and Xiang Bai.
\newblock Query-based temporal fusion with explicit motion for {3D} object
  detection.
\newblock In \emph{Thirty-seventh Conference on Neural Information Processing
  Systems}, 2023.
\newblock URL \url{https://openreview.net/forum?id=gySmwdmVDF}.

\bibitem[Hou et~al.(2024)Hou, Wang, Ye, Liu, Gong, Tan, Ding, Wang, and
  Bai]{open}
Jinghua Hou, Tong Wang, Xiaoqing Ye, Zhe Liu, Shi Gong, Xiao Tan, Errui Ding,
  Jingdong Wang, and Xiang Bai.
\newblock {OPEN}: Object-wise position embedding for multi-view {3D} object
  detection.
\newblock In \emph{European Conference on Computer Vision}, pages 146--162.
  Springer, 2024.

\bibitem[Huang and Huang(2022)]{bevdet4d}
Junjie Huang and Guan Huang.
\newblock {BEVDet4D}: Exploit temporal cues in multi-camera {3D} object
  detection.
\newblock \emph{arXiv preprint arXiv:2203.17054}, 2022.

\bibitem[Huang et~al.(2021)Huang, Huang, Zhu, Ye, and Du]{bevdet}
Junjie Huang, Guan Huang, Zheng Zhu, Yun Ye, and Dalong Du.
\newblock {BEVDet}: High-performance multi-camera {3D} object detection in
  bird-eye-view.
\newblock \emph{arXiv preprint arXiv:2112.11790}, 2021.

\bibitem[Ji et~al.(2024)Ji, Liang, and Cheng]{qaf2d}
Haoxuanye Ji, Pengpeng Liang, and Erkang Cheng.
\newblock Enhancing {3D} object detection with {2D} detection-guided query
  anchors.
\newblock In \emph{Proceedings of the IEEE/CVF Conference on Computer Vision
  and Pattern Recognition}, pages 21178--21187, 2024.

\bibitem[Jiang et~al.(2024)Jiang, Li, Liu, Wang, Jia, Wang, Han, and
  Zhang]{far3d}
Xiaohui Jiang, Shuailin Li, Yingfei Liu, Shihao Wang, Fan Jia, Tiancai Wang,
  Lijin Han, and Xiangyu Zhang.
\newblock {Far3D}: Expanding the horizon for surround-view {3D} object
  detection.
\newblock In \emph{Proceedings of the AAAI Conference on Artificial
  Intelligence}, volume~38, pages 2561--2569, 2024.

\bibitem[Jiang et~al.(2023)Jiang, Zhang, Miao, Zhu, Gao, Hu, and
  Jiang]{polarformer}
Yanqin Jiang, Li~Zhang, Zhenwei Miao, Xiatian Zhu, Jin Gao, Weiming Hu, and
  Yu-Gang Jiang.
\newblock {PolarFormer}: Multi-camera {3D} object detection with polar
  transformer.
\newblock In \emph{Proceedings of the AAAI Conference on Artificial
  Intelligence}, volume~37, pages 1042--1050, 2023.

\bibitem[Koh et~al.(2023)Koh, Lee, Lee, Kim, and Choi]{mgtanet}
Junho Koh, Junhyung Lee, Youngwoo Lee, Jaekyum Kim, and Jun~Won Choi.
\newblock {MGTANet}: Encoding sequential {LiDAR} points using long short-term
  motion-guided temporal attention for {3D} object detection.
\newblock In \emph{Proceedings of the AAAI Conference on Artificial
  Intelligence}, volume~37, pages 1179--1187, 2023.

\bibitem[Lee et~al.(2019)Lee, Hwang, Lee, Bae, and Park]{vovnet}
Youngwan Lee, Joong-won Hwang, Sangrok Lee, Yuseok Bae, and Jongyoul Park.
\newblock An energy and {GPU}-computation efficient backbone network for
  real-time object detection.
\newblock In \emph{Proceedings of the IEEE/CVF Conference on Computer Vision
  and Pattern Recognition Workshops}, pages 0--0, 2019.

\bibitem[Li et~al.(2022)Li, Zhang, Liu, Guo, Ni, and Zhang]{dndetr}
Feng Li, Hao Zhang, Shilong Liu, Jian Guo, Lionel~M Ni, and Lei Zhang.
\newblock {DN-DETR}: Accelerate {DETR} training by introducing query denoising.
\newblock In \emph{Proceedings of the IEEE/CVF Conference on Computer Vision
  and Pattern Recognition}, pages 13619--13627, 2022.

\bibitem[Li et~al.(2023{\natexlab{a}})Li, Bao, Ge, Yang, Sun, and
  Li]{bevstereo}
Yinhao Li, Han Bao, Zheng Ge, Jinrong Yang, Jianjian Sun, and Zeming Li.
\newblock {BEVStereo}: Enhancing depth estimation in multi-view {3D} object
  detection with temporal stereo.
\newblock In \emph{Proceedings of the AAAI Conference on Artificial
  Intelligence}, volume~37, pages 1486--1494, 2023{\natexlab{a}}.

\bibitem[Li et~al.(2023{\natexlab{b}})Li, Ge, Yu, Yang, Wang, Shi, Sun, and
  Li]{bevdepth}
Yinhao Li, Zheng Ge, Guanyi Yu, Jinrong Yang, Zengran Wang, Yukang Shi,
  Jianjian Sun, and Zeming Li.
\newblock {BEVDepth}: Acquisition of reliable depth for multi-view {3D} object
  detection.
\newblock In \emph{Proceedings of the AAAI Conference on Artificial
  Intelligence}, volume~37, pages 1477--1485, 2023{\natexlab{b}}.

\bibitem[Li et~al.(2024{\natexlab{a}})Li, Lan, Alvarez, and Wu]{bevnext}
Zhenxin Li, Shiyi Lan, Jose~M Alvarez, and Zuxuan Wu.
\newblock {BEVNeXt}: Reviving dense {BEV} frameworks for {3D} object detection.
\newblock In \emph{Proceedings of the IEEE/CVF Conference on Computer Vision
  and Pattern Recognition}, pages 20113--20123, 2024{\natexlab{a}}.

\bibitem[Li et~al.(2024{\natexlab{b}})Li, Wang, Li, Xie, Sima, Lu, Yu, and
  Dai]{bevformer}
Zhiqi Li, Wenhai Wang, Hongyang Li, Enze Xie, Chonghao Sima, Tong Lu, Qiao Yu,
  and Jifeng Dai.
\newblock {BEVFormer}: learning bird’s-eye-view representation from
  {LiDAR}-camera via spatiotemporal transformers.
\newblock \emph{IEEE Transactions on Pattern Analysis and Machine
  Intelligence}, 47\penalty0 (3):\penalty0 2020--2036, 2024{\natexlab{b}}.

\bibitem[Lin et~al.(2017)Lin, Doll{\'a}r, Girshick, He, Hariharan, and
  Belongie]{fpn}
Tsung-Yi Lin, Piotr Doll{\'a}r, Ross Girshick, Kaiming He, Bharath Hariharan,
  and Serge Belongie.
\newblock Feature pyramid networks for object detection.
\newblock In \emph{Proceedings of the IEEE Conference on Computer Vision and
  Pattern Recognition}, pages 2117--2125, 2017.

\bibitem[Lin et~al.(2023{\natexlab{a}})Lin, Lin, Pei, Huang, and
  Su]{sparse4dv2}
Xuewu Lin, Tianwei Lin, Zixiang Pei, Lichao Huang, and Zhizhong Su.
\newblock {Sparse4D} v2: Recurrent temporal fusion with sparse model.
\newblock \emph{arXiv preprint arXiv:2305.14018}, 2023{\natexlab{a}}.

\bibitem[Lin et~al.(2023{\natexlab{b}})Lin, Pei, Lin, Huang, and
  Su]{sparse4dv3}
Xuewu Lin, Zixiang Pei, Tianwei Lin, Lichao Huang, and Zhizhong Su.
\newblock {Sparse4D} v3: Advancing end-to-end {3D} detection and tracking.
\newblock \emph{arXiv preprint arXiv:2311.11722}, 2023{\natexlab{b}}.

\bibitem[Liu et~al.(2023{\natexlab{a}})Liu, Teng, Lu, Wang, and
  Wang]{sparsebev}
Haisong Liu, Yao Teng, Tao Lu, Haiguang Wang, and Limin Wang.
\newblock {SparseBEV}: High-performance sparse {3D} object detection from
  multi-camera videos.
\newblock In \emph{Proceedings of the IEEE/CVF International Conference on
  Computer Vision}, pages 18580--18590, 2023{\natexlab{a}}.

\bibitem[Liu et~al.(2022{\natexlab{a}})Liu, Li, Zhang, Yang, Qi, Su, Zhu, and
  Zhang]{dabdetr}
Shilong Liu, Feng Li, Hao Zhang, Xiao Yang, Xianbiao Qi, Hang Su, Jun Zhu, and
  Lei Zhang.
\newblock {DAB}-{DETR}: Dynamic anchor boxes are better queries for {DETR}.
\newblock In \emph{International Conference on Learning Representations},
  2022{\natexlab{a}}.
\newblock URL \url{https://openreview.net/forum?id=oMI9PjOb9Jl}.

\bibitem[Liu et~al.(2022{\natexlab{b}})Liu, Wang, Zhang, and Sun]{petr}
Yingfei Liu, Tiancai Wang, Xiangyu Zhang, and Jian Sun.
\newblock {PETR}: Position embedding transformation for multi-view {3D} object
  detection.
\newblock In \emph{European Conference on Computer Vision}, pages 531--548.
  Springer, 2022{\natexlab{b}}.

\bibitem[Liu et~al.(2023{\natexlab{b}})Liu, Yan, Jia, Li, Gao, Wang, and
  Zhang]{petrv2}
Yingfei Liu, Junjie Yan, Fan Jia, Shuailin Li, Aqi Gao, Tiancai Wang, and
  Xiangyu Zhang.
\newblock {PETRv2}: A unified framework for {3D} perception from multi-camera
  images.
\newblock In \emph{Proceedings of the IEEE/CVF International Conference on
  Computer Vision}, pages 3262--3272, 2023{\natexlab{b}}.

\bibitem[Liu et~al.(2020)Liu, Wu, and T{\'o}th]{smoke}
Zechen Liu, Zizhang Wu, and Roland T{\'o}th.
\newblock {SMOKE}: Single-stage monocular {3D} object detection via keypoint
  estimation.
\newblock In \emph{Proceedings of the IEEE/CVF Conference on Computer Vision
  and Pattern Recognition Workshops}, pages 996--997, 2020.

\bibitem[Loshchilov and Hutter(2017)]{adamw}
Ilya Loshchilov and Frank Hutter.
\newblock Decoupled weight decay regularization.
\newblock \emph{arXiv preprint arXiv:1711.05101}, 2017.

\bibitem[Meng et~al.(2021)Meng, Chen, Fan, Zeng, Li, Yuan, Sun, and
  Wang]{conditionaldetr}
Depu Meng, Xiaokang Chen, Zejia Fan, Gang Zeng, Houqiang Li, Yuhui Yuan, Lei
  Sun, and Jingdong Wang.
\newblock {Conditional DETR} for fast training convergence.
\newblock In \emph{Proceedings of the IEEE/CVF International Conference on
  Computer Vision}, pages 3651--3660, 2021.

\bibitem[Park et~al.(2022)Park, Xu, Yang, Keutzer, Kitani, Tomizuka, and
  Zhan]{solofusion}
Jinhyung Park, Chenfeng Xu, Shijia Yang, Kurt Keutzer, Kris Kitani, Masayoshi
  Tomizuka, and Wei Zhan.
\newblock Time will tell: New outlooks and a baseline for temporal multi-view
  {3D} object detection.
\newblock \emph{arXiv preprint arXiv:2210.02443}, 2022.

\bibitem[Philion and Fidler(2020)]{lss}
Jonah Philion and Sanja Fidler.
\newblock Lift, splat, shoot: Encoding images from arbitrary camera rigs by
  implicitly unprojecting to {3D}.
\newblock In \emph{European Conference on Computer Vision}, pages 194--210.
  Springer, 2020.

\bibitem[Vaswani et~al.(2017)Vaswani, Shazeer, Parmar, Uszkoreit, Jones, Gomez,
  Kaiser, and Polosukhin]{attention}
Ashish Vaswani, Noam Shazeer, Niki Parmar, Jakob Uszkoreit, Llion Jones,
  Aidan~N Gomez, {\L}ukasz Kaiser, and Illia Polosukhin.
\newblock Attention is all you need.
\newblock \emph{Advances in Neural Information Processing Systems}, 30, 2017.

\bibitem[Wang et~al.(2023{\natexlab{a}})Wang, Jiang, and Li]{focalpetr}
Shihao Wang, Xiaohui Jiang, and Ying Li.
\newblock {Focal-PETR}: Embracing foreground for efficient multi-camera {3D}
  object detection.
\newblock \emph{IEEE Transactions on Intelligent Vehicles}, 9\penalty0
  (1):\penalty0 1481--1489, 2023{\natexlab{a}}.

\bibitem[Wang et~al.(2023{\natexlab{b}})Wang, Liu, Wang, Li, and
  Zhang]{streampetr}
Shihao Wang, Yingfei Liu, Tiancai Wang, Ying Li, and Xiangyu Zhang.
\newblock Exploring object-centric temporal modeling for efficient multi-view
  {3D} object detection.
\newblock In \emph{Proceedings of the IEEE/CVF International Conference on
  Computer Vision}, pages 3621--3631, 2023{\natexlab{b}}.

\bibitem[Wang et~al.(2021)Wang, Zhu, Pang, and Lin]{fcos3d}
Tai Wang, Xinge Zhu, Jiangmiao Pang, and Dahua Lin.
\newblock {FCOS3D}: Fully convolutional one-stage monocular {3D} object
  detection.
\newblock In \emph{Proceedings of the IEEE/CVF International Conference on
  Computer Vision}, pages 913--922, 2021.

\bibitem[Wang et~al.(2022{\natexlab{a}})Wang, Zhang, Yang, and Sun]{anchordetr}
Yingming Wang, Xiangyu Zhang, Tong Yang, and Jian Sun.
\newblock {Anchor DETR}: Query design for transformer-based detector.
\newblock In \emph{Proceedings of the AAAI Conference on Artificial
  Intelligence}, volume~36, pages 2567--2575, 2022{\natexlab{a}}.

\bibitem[Wang et~al.(2022{\natexlab{b}})Wang, Guizilini, Zhang, Wang, Zhao, and
  Solomon]{detr3d}
Yue Wang, Vitor~Campagnolo Guizilini, Tianyuan Zhang, Yilun Wang, Hang Zhao,
  and Justin Solomon.
\newblock {DETR3D}: {3D} object detection from multi-view images via {3D-to-2D}
  queries.
\newblock In \emph{Conference on Robot Learning}, pages 180--191. PMLR,
  2022{\natexlab{b}}.

\bibitem[Wang et~al.(2023{\natexlab{c}})Wang, Huang, Fu, Wang, and Liu]{mv2d}
Zitian Wang, Zehao Huang, Jiahui Fu, Naiyan Wang, and Si~Liu.
\newblock Object as query: Lifting any {2D} object detector to {3D} detection.
\newblock In \emph{Proceedings of the IEEE/CVF International Conference on
  Computer Vision (ICCV)}, pages 3791--3800, October 2023{\natexlab{c}}.

\bibitem[Yang et~al.(2023)Yang, Chen, Tian, Tao, Zhu, Zhang, Huang, Li, Qiao,
  Lu, et~al.]{bevformerv2}
Chenyu Yang, Yuntao Chen, Hao Tian, Chenxin Tao, Xizhou Zhu, Zhaoxiang Zhang,
  Gao Huang, Hongyang Li, Yu~Qiao, Lewei Lu, et~al.
\newblock {BEVFormer} v2: Adapting modern image backbones to bird's-eye-view
  recognition via perspective supervision.
\newblock In \emph{Proceedings of the IEEE/CVF Conference on Computer Vision
  and Pattern Recognition}, pages 17830--17839, 2023.

\bibitem[Zhang et~al.(2022)Zhang, Li, Liu, Zhang, Su, Zhu, Ni, and Shum]{dino}
Hao Zhang, Feng Li, Shilong Liu, Lei Zhang, Hang Su, Jun Zhu, Lionel~M Ni, and
  Heung-Yeung Shum.
\newblock {DINO}: {DETR} with improved denoising anchor boxes for end-to-end
  object detection.
\newblock \emph{arXiv preprint arXiv:2203.03605}, 2022.

\bibitem[Zhang et~al.(2023)Zhang, Song, Liu, and Li]{decoupleddetr}
Manyuan Zhang, Guanglu Song, Yu~Liu, and Hongsheng Li.
\newblock {Decoupled DETR}: Spatially disentangling localization and
  classification for improved end-to-end object detection.
\newblock In \emph{Proceedings of the IEEE/CVF International Conference on
  Computer Vision}, pages 6601--6610, 2023.

\bibitem[Zhu et~al.(2019)Zhu, Jiang, Zhou, Li, and Yu]{cbgs}
Benjin Zhu, Zhengkai Jiang, Xiangxin Zhou, Zeming Li, and Gang Yu.
\newblock Class-balanced grouping and sampling for point cloud {3D} object
  detection.
\newblock \emph{arXiv preprint arXiv:1908.09492}, 2019.

\bibitem[Zhu et~al.(2021)Zhu, Su, Lu, Li, Wang, and Dai]{deformabledetr}
Xizhou Zhu, Weijie Su, Lewei Lu, Bin Li, Xiaogang Wang, and Jifeng Dai.
\newblock {Deformable DETR}: Deformable transformers for end-to-end object
  detection.
\newblock In \emph{International Conference on Learning Representations}, 2021.
\newblock URL \url{https://openreview.net/forum?id=gZ9hCDWe6ke}.

\end{thebibliography}
